\documentclass{bmvc2k}
\usepackage{capt-of}
\title{V3GAN: Decomposing Background, Foreground and Motion for Video Generation}

\addauthor{Arti Keshari}{cs19s008@cse.iitm.ac.in}{1}
\addauthor{Sonam Gupta}{cs18d005@cse.iitm.ac.in}{1}
\addauthor{Sukhendu Das}{sdas@iitm.ac.in}{1}

\addinstitution{
Visualization and Perception Lab\\
Dept. of CS\&E \\
Indian Institute of Technology, Madras\\
Chennai, India
}

\runninghead{Arti Keshari, Sonam Gupta, Sukhendu Das}{V3GAN}

\begin{document}

\maketitle

\begin{abstract}
Video generation is a challenging task that requires modeling plausible spatial and temporal dynamics in a video. Inspired by how humans perceive a video by grouping a scene into moving and stationary components, we propose a method that decomposes the task of video generation into the synthesis of foreground, background and motion. Foreground and background together describe the appearance, whereas motion specifies how the foreground moves in a video over time. We propose V3GAN, a novel three-branch generative adversarial network where two branches model foreground and background information, while the third branch models the temporal information without any supervision. The foreground branch is augmented with our novel feature-level masking layer that aids in learning an accurate mask for foreground and background separation. To encourage motion consistency, we further propose a shuffling loss for the video discriminator. Extensive quantitative and qualitative analysis on synthetic as well as real-world benchmark datasets demonstrates that V3GAN outperforms the state-of-the-art methods by a significant margin.
\end{abstract}

%-------------------------------------------------------------------------

%% make sure to mention the applications as video representation learning and 
\section{Introduction}
\label{sec:intro}
Unsupervised feature representation learning from unlabeled data has been a problem of great interest in Computer Vision. Other tasks like classification, clustering, etc. can benefit from the knowledge of content and dynamics present in the learned feature representation. Deep Generative models have achieved great success in unsupervised learning by generating images \cite{brock2018large, goodfellow2014generative, karras2017progressive, radford2015unsupervised} from latent noise vectors. In contrast, a similar level of success has not yet been achieved in video generation. This is primarily because the video data is more complex due to the presence of the temporal dimension. For generating photorealistic videos, the model must learn the abstraction of different objects along with the evolution of their motion over time.

\begin{figure}
\includegraphics[scale=0.56]{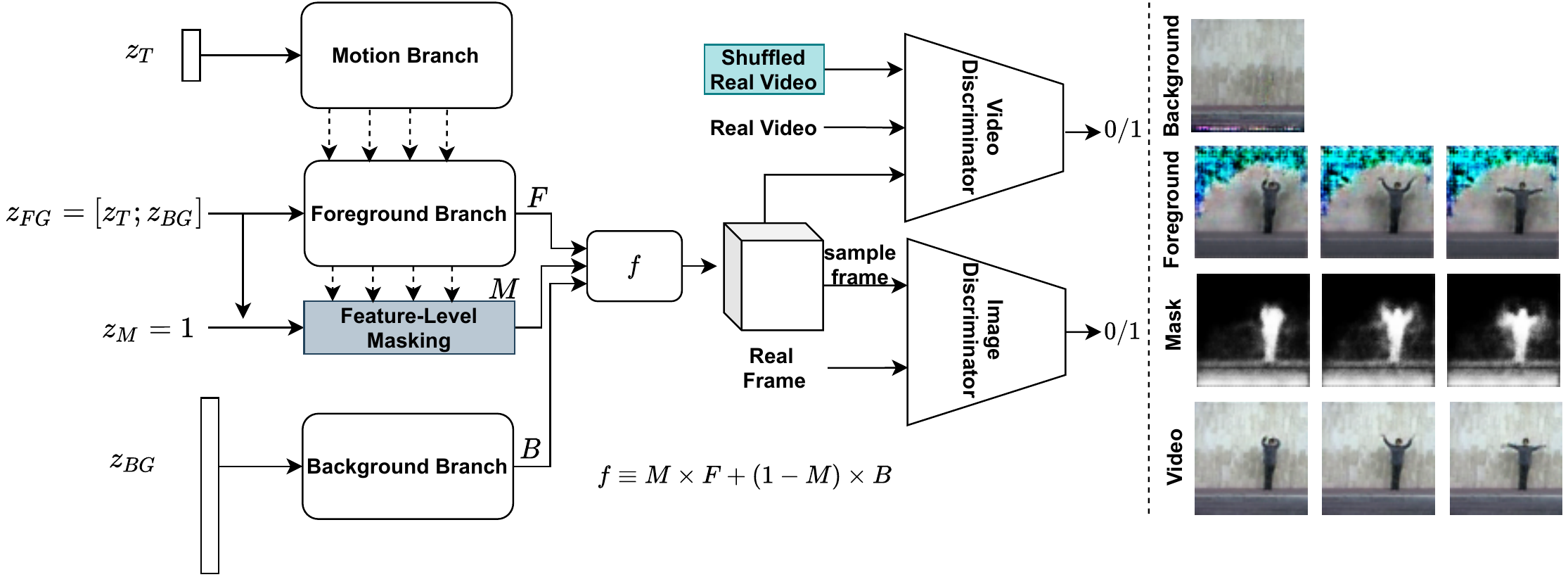}
\caption{Left: Overview of proposed V3GAN architecture. Right: Illustration of background and foreground, the mask estimated, and the final generated video frames obtained by combining the three former components.}
\label{fig:arch_overview}
\end{figure}

Many of the existing works \cite{vondrick2016generating, saito2017temporal, tulyakov2018mocogan, wang2020g3an} have proposed Generative adversarial networks (GANs) to address this task. However, the videos generated by these models are still subpar from the real videos, specifically for complex datasets like UCF101. VGAN \cite{vondrick2016generating} decomposed the task of video generation into foreground and background, but it lacked motion consistency. MoCoGAN \cite{tulyakov2018mocogan} and G3AN \cite{wang2020g3an} disentangled motion and appearance in the latent space but suffered in visual quality because they focus on the foreground and background together. Studies have shown that infants learn the physical dynamics in an unsupervised way by connecting moving things as a single object and things moving separately from one another as multiple objects \cite{spelke1990principles}. We utilize this idea and propose a novel generative method, V3GAN, which divides the task of video generation into foreground, background and motion generation, as illustrated in figure \ref{fig:arch_overview} (left). Here, the topmost branch generates the motion; the middle and bottom branches learn to separate the foreground and background using the notion of a mask. We argue that the content of a video can be described using these three key components. The foreground provides information about the main object(s) in the video, the background informs about where they are, and the motion says what they are doing. 

To maintain the frame quality and motion consistency in the generated video, similar to existing works, we also use image and video discriminators, as shown in figure \ref{fig:arch_overview} (left). Despite using these, motion across the frames of a video is not smooth. These inconsistencies can be reduced if the video discriminator can pay attention to the temporal information. To enforce this, we propose a shuffling loss where the discriminator tries to distinguish between the real video and shuffled video.  We empirically demonstrate the efficiency of the proposed method by evaluating our method on a synthetic dataset (Shapes) and real-world datasets (Weizmann Action, UCF101).  To summarize, the key contributions of the proposed method are as follows:(i) a novel framework V3GAN, which maps latent noise vectors to the background, foreground and motion for video generation. (ii) a novel feature-level masking layer that learns the mask for intermediate convolution features to obtain a refined foreground mask. (iii) a novel shuffling loss for video discriminator which complements the 3D convolution-based video discriminator by penalizing the incorrect order of frames irrespective of how realistic the individual frames are.

\section{Related Work}
Limited work has been done in video generation problems because of high computation requirements and enormous possibilities of variations in a video. A generated video may vary in content, speed, color, and intensity, but it has to be highly correlated along the temporal dimension. To tackle this problem, existing works have tried VAEs~\cite{kingma2013auto} and GAN~\cite{goodfellow2014generative} based architectures. To reduce the complexity of the video generation task, many existing approaches have focused on conditional video generation. For instance, the tasks of future frame prediction \cite{mathieu2015deep, kwon2019predicting, yu2019efficient}, video generation from single image \cite{li2018flow}, video interpolation \cite{bao2019depth, liu2017video} are all conditioned on images. Presence of appearance and structure information make these tasks simpler from unconditional video generation where the input is a noise vector. Some of these methods also use additional cues like human keypoints \cite{jang2018video, chan2019everybody}, optical flow \cite{li2018flow} etc. learned implicitly or explicitly from the input data.

\textbf{Unconditional Video Generation:}
Early video generation approaches \cite{vondrick2016generating} use basic 3D spatio-temporal convolution networks to capture spatial as well as temporal information. VGAN \cite{vondrick2016generating} attempted to disentangle the background and foreground with two-stream 3D convolution architecture. TGAN \cite{saito2017temporal} proposed to use two generators. Temporal generator outputs a sequence of noise vectors corresponding to the frames in a video. The image generator then maps these noise vectors to the RGB frames. MoCoGAN \cite{tulyakov2018mocogan} extends TGAN by replacing the temporal generator with the recurrent GRU network. It further introduced another noise called content noise and tried to disentangle the content (appearance) and motion in a video. Learning motion information in latent space makes the video inconsistent temporally. Few works \cite{Ehrhardt2020RELATEPP, Henderson2020UnsupervisedOV} have also tried object-centric approaches. Follow-up work of \cite{tulyakov2018mocogan}, G3AN \cite{wang2020g3an} proposed a three-stream generator which takes two noise vectors for motion and appearance, respectively. Third stream takes the concatenation of both noise vectors and passes it through spatio-temporal network to generate the output video. Disentangling the appearance and motion is difficult for complex real-world data like UCF101. Thus, methods like MoCoGAN, G3AN fail to generate high-quality videos for such distribution, see fig. \ref{fig:qual}.

We observe that decomposing the video generation into foreground, background and motion can significantly reduce the learning complexity, as the model now focuses on learning the temporal dynamics only for the moving object. We propose a framework that generates foreground, background and motion simultaneously to create high-quality diverse videos from the underlying data distribution. Closest to our work in the literature is VGAN~\cite{vondrick2016generating}. But unlike VGAN which shares the weight of foreground and mask generation network except the last layer, our model decomposes foreground and background at the feature level as well and propagates the mask to the next layer. This enables the network to generate good quality foreground masks. 

Several works have used the shuffle based self-supervision~\cite{misra2016shuffle, wang2015unsupervised, wang2019order} on videos as it does not require manual annotation. In \cite{wang2015unsupervised}, Wang and Gupta proposed a Siamese network \cite{misra2016shuffle} trained to sort the input sequence in correct order. Wang et al. in \cite{wang2019order} proposed a shuffle discriminator which learned whether the input optical flow maps generated by the model are shuffled or not. Unlike \cite{wang2019order}, we propose a shuffling loss which is applied to the raw RGB images from the training data. Shuffled real videos helps the discriminator to focus on temporal information even if the individual frames are realistic.
\section{Proposed Method}
We propose a three-branch deep generative model that addresses the problem of video generation. The model itself splits the problem into three sub-tasks, namely, foreground generation along with the foreground mask, background generation, and motion modelling. For learning both spatial and spatio-temporal dynamics, the model uses a frame discriminator and a video discriminator. We propose to use shuffling loss so that even small temporal inconsistencies in the model are penalized. In section \ref{generator}, we will elaborate upon the Generator of V3GAN framework along with the proposed feature level masking layer. In section \ref{discriminator}, we discuss about the discriminator. In Section \ref{shuffle_loss}, we present the proposed shuffling loss and the strategy used to generate the shuffled video. Lastly, in Section \ref{loss}, we discuss the objective function used for training V3GAN.      
\subsection{Generator}
\label{generator}
\begin{figure}
    \centering
    \includegraphics[width=\linewidth]{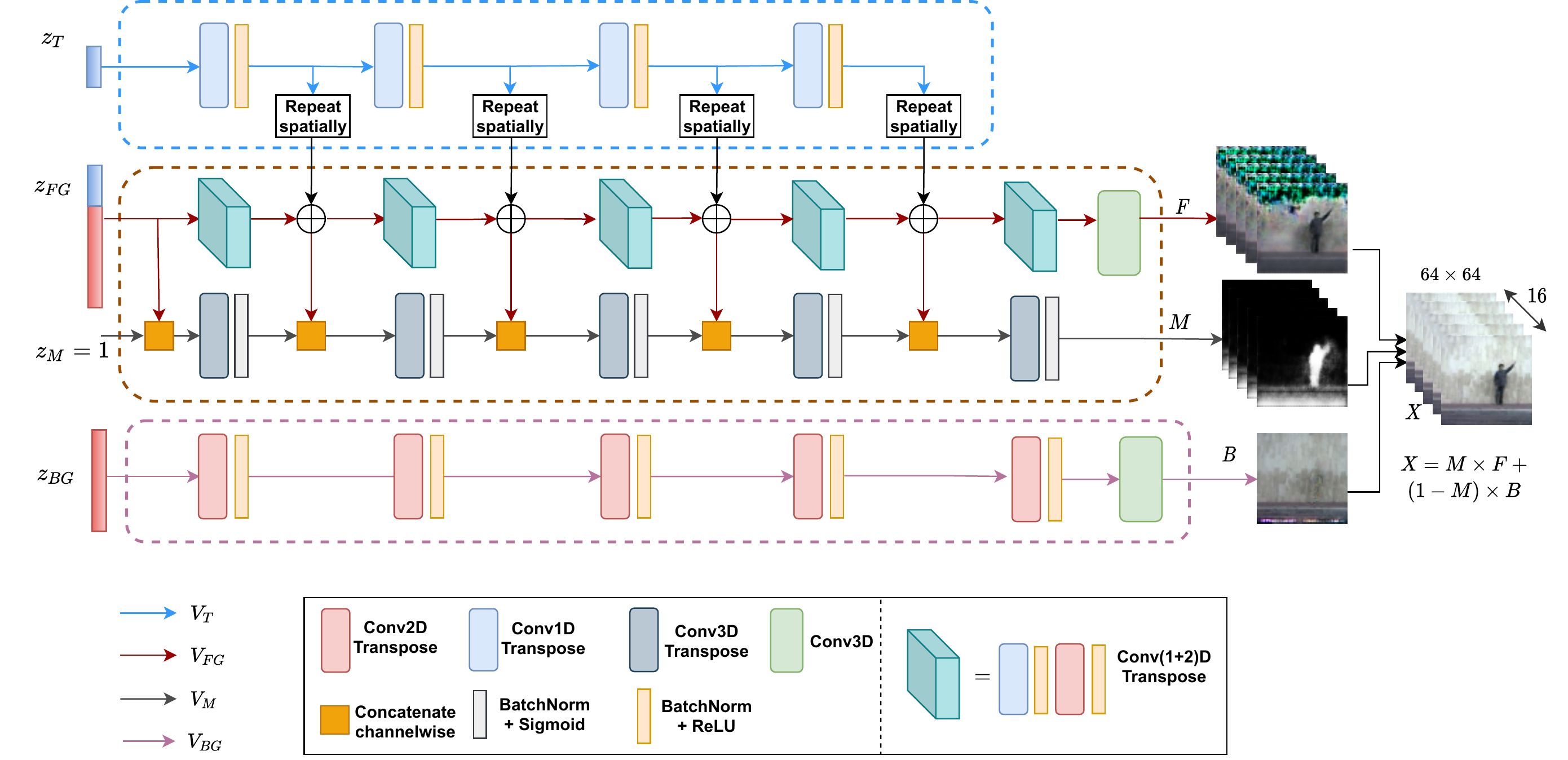}
    \caption{Architecture diagram of proposed V3GAN. $V_T$,  $V_{FG}$, $V_{BG}$ correspond to temporal, foreground, and background branches shown in blue, red, pink respectively. $V_M$ is the feature level masking layer shown in black. Input $z_T$ and $z_{BG}$ are sampled from the Gaussian noise.}
    \label{fig:arch}
\end{figure}
V3GAN generator consists of three branches, which are the temporal branch $V_T$, the foreground branch $V_{FG}$ augmented with the feature level masking layer $V_M$ and the background branch $V_{BG}$ as shown in figure \ref{fig:arch}. The inputs to the generator are two noise vectors $z_{BG}$ and $z_T$ corresponding to the processing pipeline of background and motion respectively. We notice that foreground is highly entangled with the motion, and all possible combinations of foreground and backgrounds are also not semantically meaningful. For instance, a surfer surfing on the road is an unrealistic scenario. Therefore, the foreground noise vector is chosen to be the concatenation of $z_{BG}$ and $z_{T}$.  

The choice of convolution operations is one of the key elements to enforce that each branch learns unique features. Given the assumption that there is no appreciable camera motion, i.e., the foreground object is the moving component, the background can be treated as a fixed $2D$ bitmap layer shared over the entire video. Hence, the background branch $V_{BG}$ is designed to be independent of other branches. $V_{BG}$ upsamples the input only spatially using the Conv2D transpose layer. The output of $V_{BG}$ is a single $2D$ image which corresponds to the background of the generated video. 

The $V_{T}$ branch is tightly coupled with the $V_{FG}$ branch, as the moving foreground will be highly correlated with the temporal dynamics. We use a noise vector $z_T$ as input which evolves in temporal dimension by upsampling the input with Conv1D transpose layers. This branch learns the global motion information and guides the foreground stream $V_{FG}$ by providing clues on various motion aspects like how to move, with what speed, direction of motion, etc. The output of each Conv1D transpose layer is repeated spatially to match the dimension of the foreground features. The resulting features are then combined  with the foreground features using element-wise addition.

Foreground branch $V_{FG}$ seeks to generate the spatio-temporal features. This branch consists of five Conv(1+2)D transpose layers. Conv(1+2)D operation \cite{wang2020imaginator} is factorized form of Conv3D transpose which is Conv1D transpose followed by Conv2D transpose. Conv1D transpose helps the foreground to learn the temporal component and Conv2D transpose helps to learn the spatial content of the foreground. The output features of Conv(1+2)D layer are added with the corresponding temporal features. The resulting features are then concatenated with the feature level masking layer as shown in figure \ref{fig:arch}. Conv3D layer is introduced at the output of $V_{FG}$ to get the desired foreground in RGB domain from the learned features. 

\begin{figure}
\includegraphics[width=\linewidth]{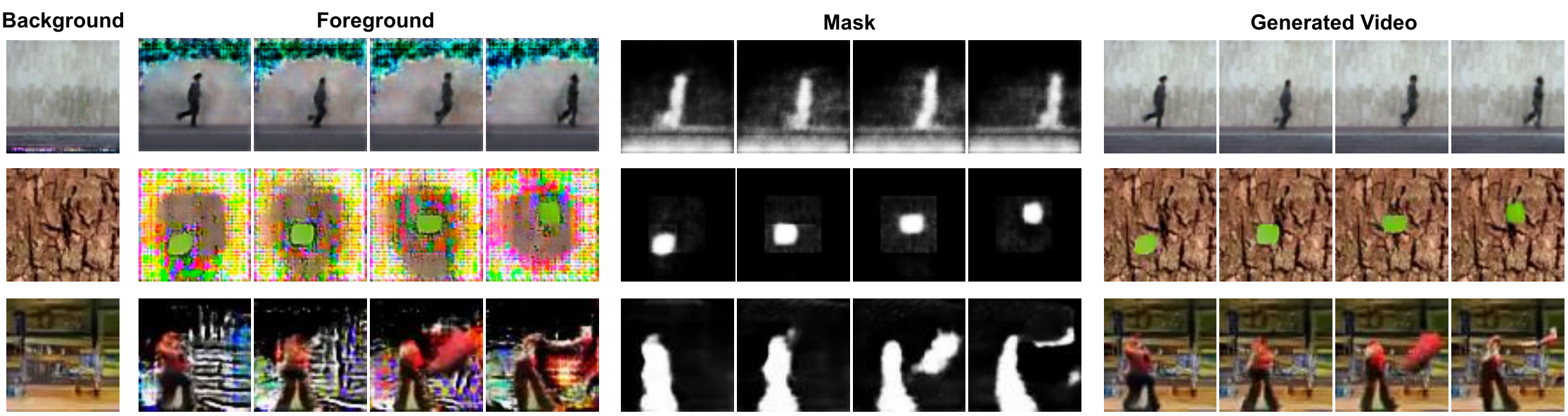}
\caption{From left to right: Illustrations of background, foreground, mask; and the final generated video obtained by combining these components on Weizmann (top), Shapes (middle), UCF101 (bottom) datasets.}
\label{fig:bg_fg_mask}
\end{figure}

\textbf{Feature level masking:}
Another key element for achieving the successful decomposition of foreground and background is the quality and accuracy of the generated foreground mask. Inspired by \cite{vondrick2016generating}, we attempt to learn the mask from the foreground branch $V_{FG}$ in an unsupervised manner by using the following relationship between the foreground ($F$), background ($B$) and mask ($M$) for video $X$. 
\begin{equation}
    X_i = M_i \times F_{i} + (1 - M_i) \times B_{i}, \quad \quad M_i \in [0,1] 
\end{equation}
where $X_i$, $M_i$, $F_i$ and $B_i$ are the $i^{th}$ pixel in the video, mask, foreground and background respectively.

The input to the masking layer is a fixed vector $z_M = 1$ concatenated with $z_{FG}$ along the channel dimension.
Considering $z_M = 1$ implies that $z_{FG}$ completely belongs to the foreground, which enforces the branch $V_{FG}$ to focus on foreground features. The implicit assumption here is that at the initial layer, all channels of $z_{FG}$ represent the foreground. The input is then passed through five Conv3D transpose layers to learn the spatio-temporal features. Each intermediate masking feature is concatenated with foreground features. Output of each stage has a single channel with sigmoid activation function. Empirical results show that feature level masking improves the performance of the video generation substantially. The evaluation results are reported later in section \ref{exps}.

\subsection{Discriminator}
\label{discriminator}
V3GAN consists of two discriminators similar to MoCoGAN \cite{tulyakov2018mocogan}, a video discriminator and an image discriminator. Unlike MoCoGAN, we also pass shuffled video as input to video discriminator. Video discriminator contains Conv(2+1)D layers and the image discriminator contains Conv2D layers. We used spectral normalization \cite{miyato2018spectral} in the discriminator as well as in generator to stabilize the training. 

\subsection{Shuffling Loss}
\label{shuffle_loss}
The video discriminator $D_V$ uses (2+1)D convolution to focus on both spatial and temporal aspects of the input. We propose a shuffling loss that exploits the sequential ordering of frames of a video. This loss enforces $D_V$ to emphasize on the correctness of motion. The knowledge of incorrect ordering of frames vs. the correct ordering of frames is imparted to $D_V$, explicitly, by training it to classify a shuffled real video sequence as fake. Let $sh$ be the shuffling function which takes the real video $X$ and shuffling parameter $\alpha$ as input to generate the shuffled video that is hard for the discriminator to classify. The shuffling loss is then defined as follows:
\begin{equation}
\label{l_shuffle}
L_{shuffle}(X, \alpha)= E[\log(1 - D_V(sh(X, \alpha))]
\end{equation}
The shuffling parameter controls the fraction of frames to be shuffled. The function $sh$ is defined as follows:\\
\textbf{Step 1:} Select $\alpha N$ frames uniformly at random from the real video where $N$ (here 16) is the length of the video. \textbf{Step 2:} Swap a selected frame with its first or second neighbouring frame.
The empirical analysis for shuffling loss can be found in Section \ref{ablation}.

\subsection{Loss Function}
\label{loss}
The feedback from both $D_V$ and $D_I$ are used to train the network. Both $D_V$ and $D_I$ use adversarial loss functions as proposed in \cite{radford2015unsupervised}.  The learning problem of V3GAN is given in eq. \ref{eq:loss_function}
where $F_I$ is the loss function of image discriminator and $F_V$ is the loss function of video discriminator. The definitions for these losses are given as:
\begin{gather}
\label{eq:loss_function}
\min_{G} \max_{D_I,D_V} (F_I(G, D_I) + F_V(G, D_V))\\
F_I = E[\log{D_I(X_f)}]+E[\log(1-D_I(G(z_{BG},z_T)_f))] \\
F_{V} = E[\log{D_V(X)}]+E[\log(1-D_V(G(z_{BG},z_T)))] + L_{shuffle}(X, \alpha)    
\end{gather}
G represents the generator. Subscript $f$ in the loss function of $D_{I}$ refers to the randomly sampled frame from the video. $L_{shuffle}$ represents the shuffling loss as described in eq. \ref{l_shuffle}.

\section{Experiments}
\label{exps}
We evaluate our method on the following datasets:

\textbf{Shapes Dataset} \cite{tulyakov2018mocogan} is a synthetic dataset, containing 4000 videos of circles and squares of different colors and sizes on a black background. To understand the ability of the model to decompose foreground, background and motion, we randomly sample a texture as background from 7 different textures and apply it to each moving shape video. 

\textbf{Weizmann Action Dataset}~\cite{gorelick2007actions} contains 93 videos of 9 people performing 10 actions, including running, jumping jack, etc. We augment the data by horizontal flipping. 

\textbf{UCF101}~\cite{soomro2012ucf101} is a commonly used dataset for video generation, containing more complex and variety of videos. It includes 13,220 videos of 101 different action categories. Since UCF101 contains some videos with moving background, we stabilize the camera motion by adopting the stabilization process as in \cite{vondrick2016generating}.
For both Action and UCF101 datasets, the video frames are rescaled to 85x64 then centre cropped to 64x64, same as in \cite{saito2017temporal}.

\textbf{Implementation Details:}
For all our experiments, the dimension of $z_{BG}$ is set to 128 and that of $z_{T}$ is set to 10. Shuffling parameter is set to $\alpha = 0.5$ and batch size is $16$. The input videos contain 16 frames with resolution of $64 \times 64$. Adam optimizer \cite{kingma2014adam} is used to train the network with $\beta_1 = 0.5$ and $\beta_2 = 0.999$. Learning rate for generator and discriminators is set to $10^{-4}$. The details of the network architecture design can be found in the supplementary document (SD). Source code will be made public.
\subsection{Quantitative Evaluation}
%Quantitative evaluation of video generation models is a difficult problem due to non-availability of corresponding ground truths.
%FID metric is the squared Wasserstein distance between two multidimensional Gaussian distributions: ${\displaystyle {\mathcal {N}}(\mu ,\Sigma )}$, in the case of Video generative GAN, the distribution are calculated by passing real video and generated video through deep network( ResNet101 in our case) and get the mean and covariance of the distribution. These mean and covariance is then used to calculate FID metric. Lower FID indicates better quality of the videos.  
 %${\displaystyle {\text{FID}}=|\mu -\mu _{w}|^{2}+\operatorname {tr} (\Sigma +\Sigma _{w}-2(\Sigma \Sigma _{w})^{1/2})}$

%   Inception score is the KL divergence between class conditional and marginal probability distribution $exp(Exp( KL(p(y|x)||p(y))))$. 
%  So, we have also used FID and IS an evaluation metric for comparing our work with previous state-of-the-art works. 
  %Inception score is the KL divergence between class conditional and marginal probability distribution $exp(Exp( KL(p(y|x)||p(y))))$. Class conditionals ensure the quality of video and marginal probability ensures diversity of the video. Higher the value of IS better is the model. 
  
  We compare the proposed method quantitatively with four state-of-the-art (SOTA) methods, VGAN, TGAN, MoCoGAN and G3AN using Frechet Inception Distance (FID)~\cite{heusel2017gans} and Inception Score(IS) \cite{salimans2016improved}. FID metric is the squared Wasserstein distance between two multidimensional Gaussian distributions: ${\displaystyle {\mathcal {N}}(\mu ,\Sigma )}$. We used a deep 3d CNN network of ~\cite{hara2018can} to extract the mean and covariance of the distributions, which is then used to calculate the FID metric using ${\displaystyle {\text{FID}}=|\mu -\mu _{w}|^{2}+\operatorname {tr} (\Sigma +\Sigma _{w}-2(\Sigma \Sigma _{w})^{1/2})}$. Lower FID indicates better quality of the generated videos. For evaluation, we generated 5000 video samples using the trained model and calculate FID value. IS is the KL divergence between class conditional and marginal probability distribution $\exp(E_{x \sim p_g}(KL(p(y|x)||p(y))))$. Since the inception model must be pretrained on the data for which IS is calculated, we reported IS only for UCF101. 
  
%   We compare the proposed method quantitatively with four state-of-the-art (SOTA) methods, VGAN, TGAN, MoCoGAN and G3AN using Frechet Inception Distance (FID)~\cite{heusel2017gans} for Shapes, Weizmann, and UCF101 datasets. FID metric is the squared Wasserstein distance between two multidimensional Gaussian distributions: ${\displaystyle {\mathcal {N}}(\mu ,\Sigma )}$. In the case of video GANs, the real and generated distributions are calculated by  extracting the pre-trained features using  deep 3d CNN network of ~\cite{hara2018can}. The mean and covariance of these distributions are then used to calculate the FID metric as ${\displaystyle {\text{FID}}=|\mu -\mu _{w}|^{2}+\operatorname {tr} (\Sigma +\Sigma _{w}-2(\Sigma \Sigma _{w})^{1/2})}$. Lower FID indicates better quality of the generated videos. For evaluation, we generated 5000 video samples using the trained model and calculate FID value. 
 
\begin{figure}
    \centering
    \includegraphics[scale=0.6]{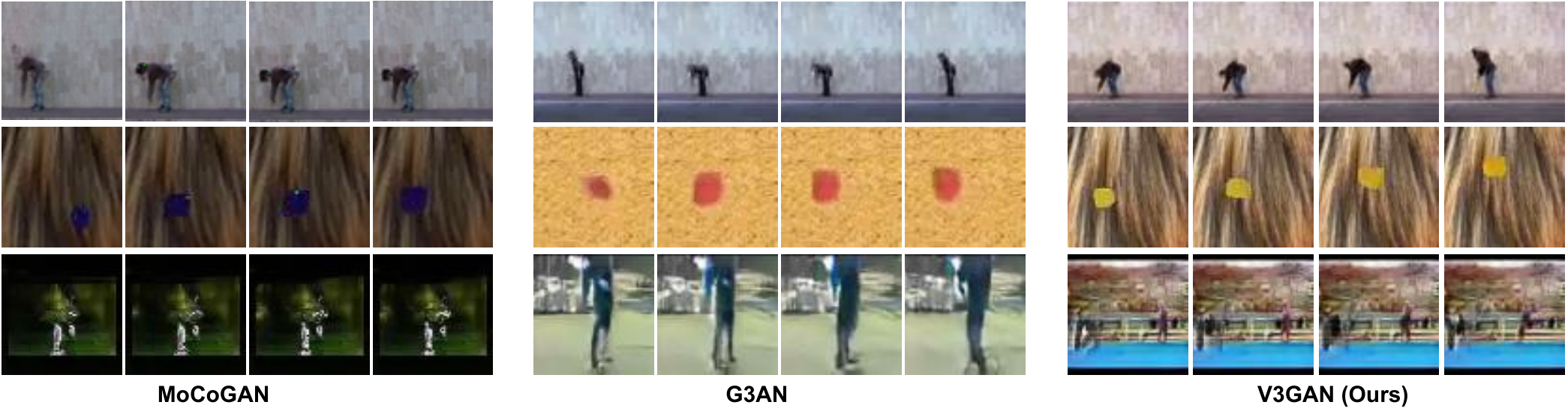}
    \caption{Qualitative comparison of performance with state-of-the-art methods: MoCoGAN and G3AN, on: Weizmann (top), Shapes (middle) and UCF101 (bottom) datasets. For the sake of illustration, frames are sampled at equal intervals from the video.}
    \label{fig:qual}
\end{figure}  

\begin{table}[]
\centering
\begin{tabular}{|l|c|c|c|c|c|c|} \hline
\textbf{Method}        & \textbf{Shapes} & \textbf{Weizmann} & \multicolumn{2}{c|}{\textbf{UCF101}} & \multicolumn{1}{c|}{\textbf{\begin{tabular}[c]{@{}c@{}}UCF101\\ (Stable)\end{tabular}}} \\ 
                       & \textbf{(FID$\downarrow$)}  & \textbf{(FID$\downarrow$)}    & \textbf{(FID$\downarrow$)}    & \textbf{(IS$\uparrow$)}    & \textbf{(FID$\downarrow$)}                             \\ \hline
VGAN [4]                 & -               & 158.04            & 115.06            & 2.94             & -                                                                                   \\
TGAN [3]                 & -               & 99.85             & 110.58            & 2.74             & -                                                                                    \\
MoCoGAN [2]             & $144.87^{\dag}$          & 92.18             & 104.14            & 3.06             & $218.59^\dag$                                                                              \\
G3AN [1]                   & -                & 86.01                    & 91.21                   & 3.62                  & -                                                                                    \\
G3AN*                  & $168^{\dag}$             & $68.19^{\dag}$             & $86.73^{\dag}$             & 3.44             & $102.13^{\dag}$                                                                            \\
V3GAN (w/o shuffle)    & 62.59           & 64.33             & \textbf{78.71}    & 3.84             & 75.64                                                                             \\
V3GAN + shuffle (Ours) & \textbf{28.07}  & \textbf{62.65}    & 80.18             & \textbf{3.88}    & \textbf{74.36}                                      \\ \hline                    
\end{tabular}
\caption{Quantitative comparison with SOTA methods using FID metric for Shapes, Action, UCF101 including stabilized UCF101 datasets. $\dag$ indicates that the values are obtained by training the official codes provided by the authors. G3AN* contains value after running official code with modified discriminator, which uses Conv(2+1)D instead of Conv3D.}
\label{tab:quantitative}
\end{table}
Table \ref{tab:quantitative} contains the FID and IS for four prior published works along with ours. Our method outperforms all other baseline methods suggesting that the learned data distribution is closer to the real data distribution. V3GAN outperforms the SOTA even without using shuffling loss indicating that the decomposed representation is able to learn the spatio-temporal dynamics better. We note that, for Shapes dataset, the FID of our method is significantly lower than G3AN model. On visualizing the generated videos using both methods, we found that the videos generated by G3AN lack diversity. It generates a combination of only 3 backgrounds out of 7, whereas V3GAN generates videos with all 7 backgrounds. Moreover, by design, V3GAN uses the same background over all frames in a video leading to a much better temporal consistency.  Since the background of synthetic data does not contain illumination, shadows and other inherent variations over frames, it gives an impetus to our model for a drastic improvement for Shapes dataset. UCF101 dataset contains small background motion. Therefore, we have trained our model on the original data as well as stabilized data. Our method outperforms SOTA methods in both scenarios. In the last row of table \ref{tab:quantitative}, we show the effect of using the proposed shuffling loss. It is evident that shuffling improves the temporal consistency of the video resulting in better FID values on all datasets.
\begin{figure}
    \centering
    \includegraphics[scale=0.58]{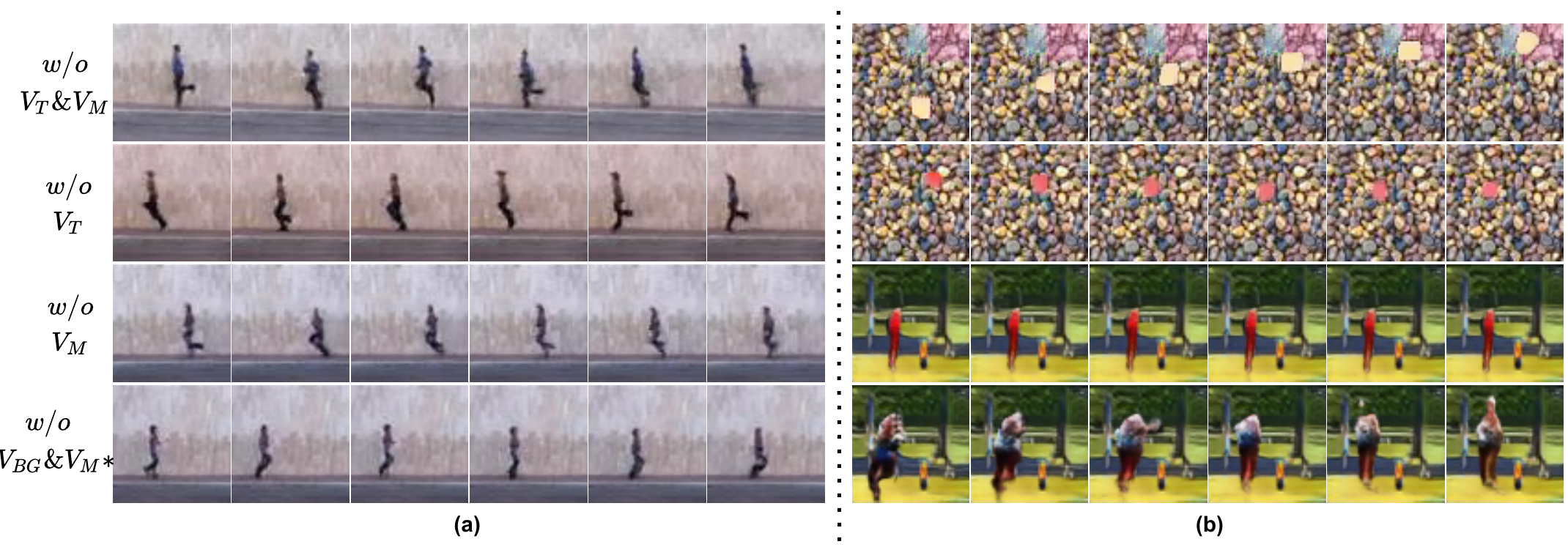}
    \caption{(a) Ablation study: Generated videos obtained by removing each branch of V3GAN. (b) Examples of videos generated by V3GAN on Shapes (first two rows and UCF101 (last two rows) datasets.}
    \label{fig:qual_ablation}
\end{figure}
\subsection{Qualitative Evaluation}
In figure \ref{fig:bg_fg_mask}, we show the background, foreground and mask generated by our model along with the generated video for the three datasets. Our network is able to decompose the foreground and background with the help of the mask without any supervision. To verify the improvement in the visual quality of the generated videos, we compare our results with that of MoCoGAN and G3AN qualitatively. Figure \ref{fig:qual} shows that, our method is able to generate in-line appearance of the front content. In the first row, the hands of the person are clearly visible, and in the second row the shape of the rectangle is maintained. In the third row, the background is of good quality. Although there is scope of improvement in the foreground, but it is modelled better than the existing methods. Figure \ref{fig:qual_ablation} (b) shows the videos generated on Shapes and UCF101 datasets by keeping the background noise fixed and only altering the foreground noise. It shows that motion pattern is highly correlated with the foreground.

\begin{figure}
    \centering
    \includegraphics[scale=0.52]{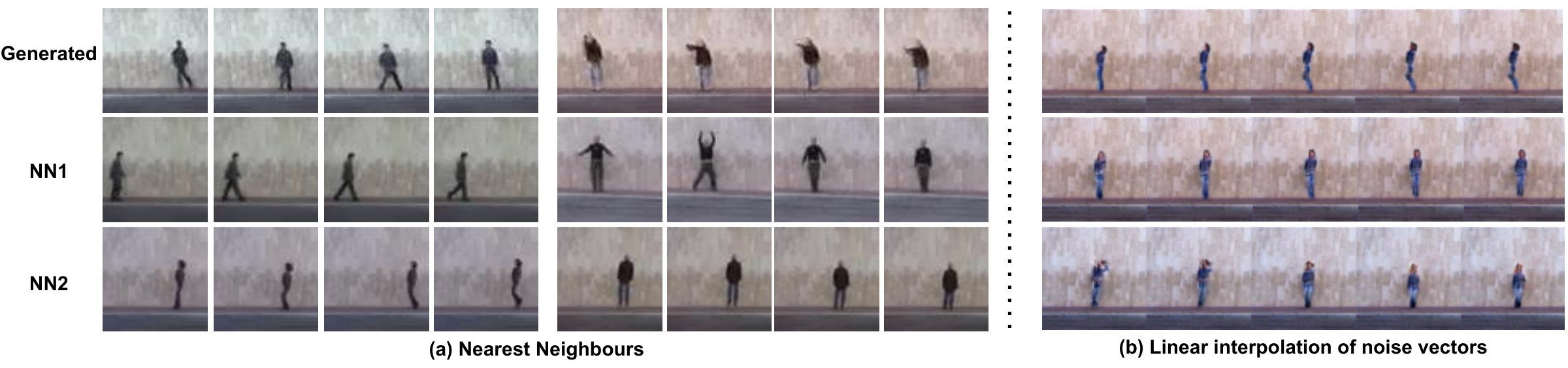}
    \caption{\textbf{(a) Nearest neighbours:} Top row contains generated video sequences. NN1 and NN2 represent first and second nearest neighbours of the generated video. \textbf{(b) Linear interpolation:} 
    Top and bottom rows represent video generated from two noise vectors. Middle row is generated from an intermediate noise vector between them.}
    \label{fig:mem}
\end{figure}

\textbf{Nearest neighbours and Linear interpolation Analysis: }
As the Weizmann dataset has a small size, we further examine whether or not the proposed method has memorized the training data using nearest neighbour and linear interpolation similar to \cite{brock2018large, saito2020train}. To find the nearest neighbours, we computed the euclidean distance between the feature embeddings of the generated videos with that of the training dataset. From figure \ref{fig:mem} (a), we observe that the first two nearest neighbours chosen from the training dataset are distinct from the generated samples. In case of interpolation, we generate video corresponding to the intermediate noise vectors between two sampled noise vectors. It can be seen from figure \ref{fig:mem} (b) that there is a graceful transition between videos with change in input noise. Above observations suggest that the network has not memorized the training dataset.

\begin{table}[]
\begin{minipage}{.6\linewidth}
\center
\begin{tabular}{|l|c|c|c|}
\hline
\textbf{Architecture} & \textbf{\begin{tabular}[c]{@{}c@{}}Shapes\\ (FID$\downarrow$)\end{tabular}} & \textbf{\begin{tabular}[c]{@{}c@{}}Weizmann\\ (FID$\downarrow$)\end{tabular}} & \textbf{\begin{tabular}[c]{@{}c@{}}UCF101\\ (FID$\downarrow$)\end{tabular}} \\
\hline
w/o $V_{T}$ \& $V_M$  & 40.93                                                                       & 73.44                                                                         & 81.60                                                                             \\
w/o $V_{BG}$ \& $V_M*$ & 168.56                                                                      & 70.9                                                                          & 198.37                                                                             \\
w/o  $V_{M}$          & 34.19                                                                       & 74.45                                                                         & 76.33                                                                             \\
w/o $V_T$             & 66.5                                                                        & 70.68                                                                         & 80.29                                                                             \\
V3GAN (Ours)          & \textbf{28.07}                                                              & \textbf{62.65}                                                                & \textbf{74.36}   \\
\hline
\end{tabular}
\caption{Effect of different components of V3GAN, studied for all three datasets. Here, $V_{M}*$ means masking has also been removed from the last layer.}
\label{tab:ablation_table}
\end{minipage} \hfill
\begin{minipage}{0.35\linewidth}
\centering
\includegraphics[scale=0.2]{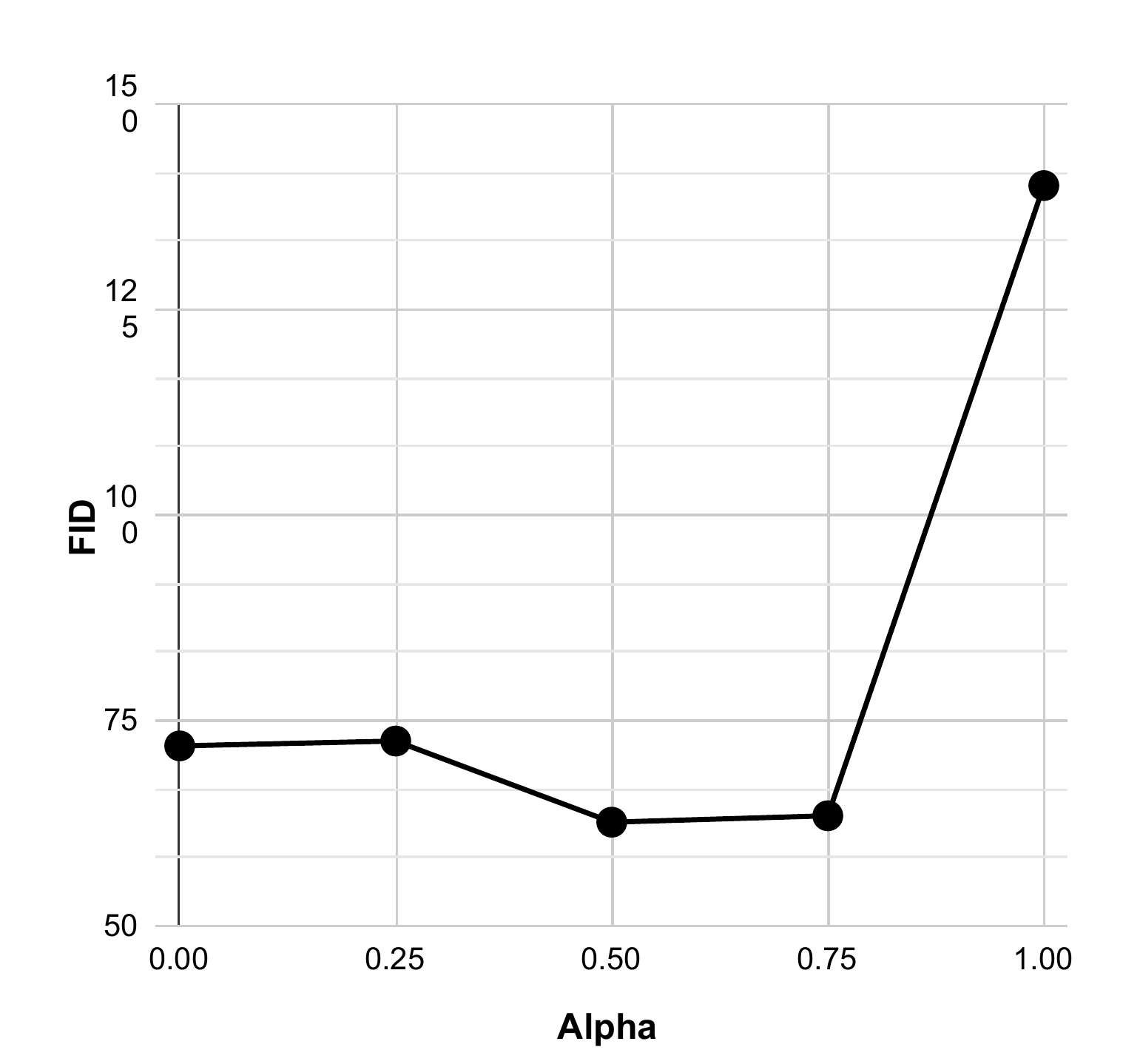}
\captionof{figure}{Effect of changing $\alpha$ in shuffling loss, on FID value for Action Dataset.}
\label{fig:ablation_shuffle}
\end{minipage}
\end{table}

\begin{figure}
    \centering
    \includegraphics[scale=0.55]{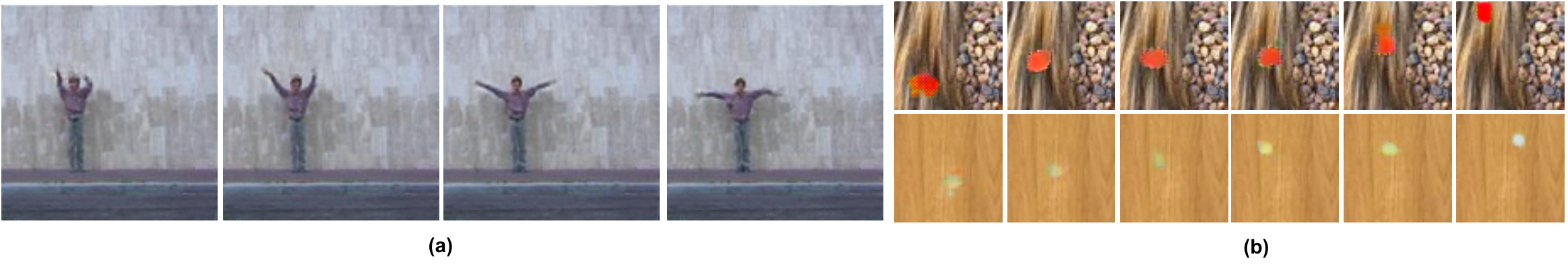}
    \caption{(a) Video generation at 128x128 resolution on Weizmann dataset. (b) Longer video sequence (32 frames) generation on Shapes dataset. Frames are chosen at regular interval for visualization purpose.}
    \label{fig:high-long}
\end{figure}

\subsection{Additional Analysis}
\label{ablation}
\textbf{Ablation study:} In table \ref{tab:ablation_table}, we show the importance of each component of our architecture by removing different branches. We first remove $V_T$ branch and $V_M$ layer while retaining the mask generation at the last layer of the model. This configuration leads to a significant drop in performance proving that both feature level masking and motion branch are important to get good quality videos. We then removed $V_{BG}$ and $V_M$ (including masking at the last layer). After this, $V_M$ and $V_T$ were removed individually. Refer SD for complete architecture. It can be observed in figure \ref{fig:qual_ablation} (a), that the quality of generated videos degraded when removing each component individually. For Weizmann dataset, removing $V_M$ leads to the worst performance, confirming the contribution of the feature level masking layer. Interestingly, for Shapes and UCF101 dataset, removing $V_{BG}$ and $V_M*$ causes the maximum drop in performance due to mode collapse.  

% Finally, we study the effect of varying the shuffling parameter($\alpha$). 

\textbf{Effect of shuffling parameter ($\alpha$):} It can be seen in figure \ref{fig:ablation_shuffle}, that the learning is hindered at both lower and higher values of $\alpha$. The best results are obtained at $\alpha=0.5$. This means that higher amount  of shuffling makes it obvious for the discriminator to identify as fake video. On the other hand, shuffling fewer frames generates difficult examples that contributes to overall learning.%This means shuffling all frames make it obvious for discriminator to identify as fake and shuffling a few frames give an opportunity to discriminate between true and shuffled sequences.

\textbf{Longer Video Generation and Higher Resolution Video Generation:} We marginally modify V3GAN to check the scalability of the proposed method. First, we added one more layer in each branch of our architecture to generate the videos at 128x128 resolution with 16 frames. The qualitative results on Weizmann dataset are demonstrated in figure \ref{fig:high-long}(a). Then, we generated longer videos of 32 frames on Weizmann and Shapes datasets. Figure \ref{fig:high-long}(b) shows the generated samples for Shapes data, while results on Weizmann dataset can be found in the SD. We found that our model can be adopted for higher resolution and longer video generation without significant modifications and parameter overhead. 

\textbf{Class Conditional Video generation:} Apart from the above experiments, we also test our model for class conditional video generation on Weizmann dataset. We append one-hot vector that corresponds to the class label with $z_T$ noise vector and find that our model can correctly generate videos for given input class. Qualitative examples can be found in SD. 

%The learnable parameters in the mentioned settings varies as (64x64, 16 frames): 21.5M, (64x64, 32 frames) : 21.7 M and (128x128, 16 frames) : 23.7 M. 

\section{Conclusion}
In this work, we have presented a generative model V3GAN, that learns to synthesize videos  from a latent Gaussian space by generating background, foreground and motion simultaneously. We have introduced a novel feature masking layer and shuffling loss. Qualitative and  quantitative evaluations on three datasets show that our method outperforms all SOTA methods. Ablation studies prove the individual contribution of each component in the architecture towards  achieving the best results. The assumption that the background is static works reasonably well, as we target to generate short video clips of only 16 frames duration. For future work, it will be an interesting direction to explore if the background and foreground can be separated irrespective of camera motion, multiple moving objects and clutter.    
\bibliography{v3gan}
\end{document}

\maketitle

\section{Overview}
 The supplementary is organized as follows. In section \ref{sec:arch}, we provide details about the network architecture of V3GAN. In Section \ref{sec:additional}, we show qualitative examples for Conditional video generation using class labels and Longer sequence (32 frames) generation for Weizmann dataset. In section \ref{sec:ablation}, we present additional ablation studies on the proposed shuffling loss. We also show the architecture diagrams of the ablation studies performed in Section 4.3 of the main paper. Lastly, in section \ref{sec:qual}, we show more samples of generated videos along with foreground, background and mask on Shapes and UCF101 datasets. We also show some failure cases on Weizmann dataset.    
 
 For rest of the supplementary, we follow the notations given below :\\
 $z_{BG}$: Background noise vector, $z_T$: Motion noise vector sampled from Standard Normal distribution, $z_{FG}$: foreground noise vector constructed by concatenating $z_{BG}$ and $z_T$. For illustration purpose, only 10 frames of each video are displayed in a row for all figures. Unconditionally generated videos from V3GAN can be found at link: \href{https://drive.google.com/drive/folders/11cRwAZBlENPA6pWzydAre5cCz0JKs4M_?usp=sharing}{examples}.

\section{Network Architecture}
\label{sec:arch}
We discuss the details of the proposed model. V3GAN generator consists of three branches, namely, motion $V_T$, foreground $V_{FG}$, and background $V_{BG}$. The foreground branch is augmented with the proposed feature-level masking layer $V_M$. V3GAN maps 1x1x1 noise vectors to a sequence of 16 RGB frames of resolution 64x64. In Table \ref{tab:arch}, we present the number of output channels along with the shape of the output features at each stage for every branch in the network. The details of the masking layer has also been shown in a separate column. 

\begin{table}[]
\centering
\begin{tabular}{|c|c|c|c|c|c|c|c|c|} \hline
\textbf{}                                    & \multicolumn{2}{c|}{\textbf{\begin{tabular}[c]{@{}c@{}}Motion\\ Branch\end{tabular}}} & \multicolumn{2}{c|}{\textbf{\begin{tabular}[c]{@{}c@{}}Foreground\\ Branch\end{tabular}}} & \multicolumn{2}{c|}{\textbf{\begin{tabular}[c]{@{}c@{}}Masking\\ Layer\end{tabular}}} & \multicolumn{2}{c|}{\textbf{\begin{tabular}[c]{@{}c@{}}Background\\ Branch\end{tabular}}} \\ \hline
\textbf{Layers}                              & \textbf{Ch}                & \multicolumn{1}{l|}{\textbf{TxHxW}}                & \textbf{Ch}                  & \multicolumn{1}{l|}{\textbf{TxHxW}}                  & \textbf{Ch}                 & \multicolumn{1}{l|}{\textbf{TxHxW}}                & \textbf{Ch}                  & \multicolumn{1}{l|}{\textbf{TxHxW}}                  \\ \hline
\multicolumn{1}{|l|}{\textbf{Input}} & 10                               & 1x1x1                                              & 138                                & 1x1x1                                                & 1                                 & 1x1x1                                              & 128                                & 1x1x1                                                \\
1                                            & 512                              & 4x1x1                                              & 512                                & 4x4x4                                                & 1                                 & 4x4x4                                              & 512                                & 1x4x4                                                \\
2                                            & 512                              & 8x1x1                                              & 512                                & 8x8x8                                                & 1                                 & 8x8x8                                              & 512                                & 1x8x8                                                \\
3                                            & 256                              & 16x1x1                                             & 256                                & 16x16x16                                             & 1                                 & 16x16x16                                           & 256                                & 1x16x16                                              \\
4                                            & 128                              & 16x1x1                                             & 128                                & 16x32x32                                             & 1                                 & 16x32x32                                           & 128                                & 1x32x32                                              \\
5                                            & -                                & -                                                  & 64                                 & 16x64x64                                             & 1                                 & 16x64x64                                           & 64                                 & 1x64x64                                              \\
\textbf{6}                                   & -                                & -                                                  & 3                                  & 16x64x64                                             & -                                 & {\color[HTML]{000000} -}                           & 3                                  & 1x64x64                    \\ \hline                         
\end{tabular}
\label{tab:arch}
\caption{Output dimension of each branch/layer in V3GAN. Ch specifies the number of output channels.}
\end{table}
\begin{figure}
    \centering
    \includegraphics[scale=0.75]{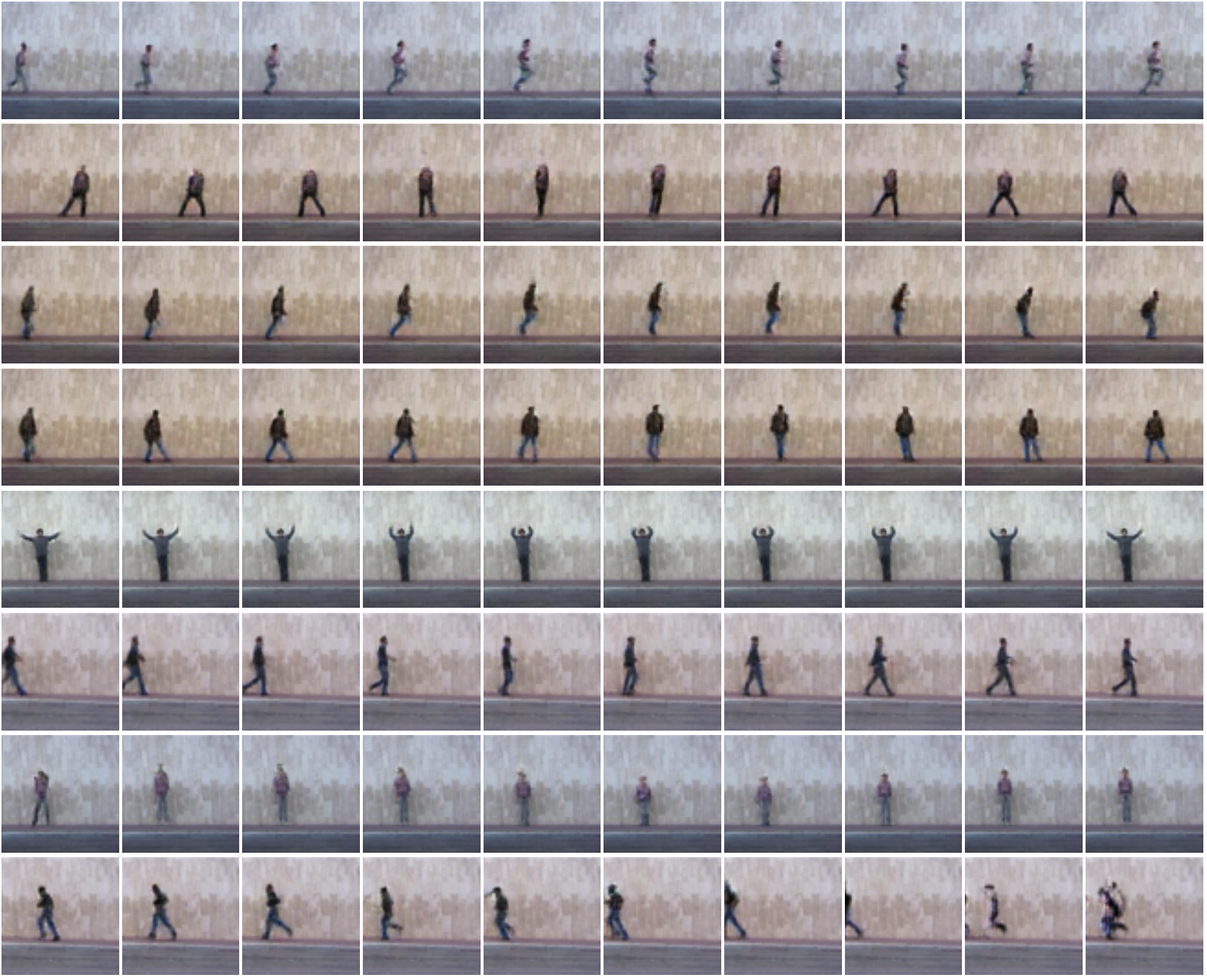}
    \caption{Unconditionally generated videos with 32 frames each. Each row represents a video sequence generated by randomly sampling $z_{M}$ and $z_{BG}$ from standard Gaussian distribution. 10 alternate frames have been picked starting from 3rd frame due to space constraint. }
    \label{fig:32frames}
\end{figure}
\section{Additional Qualitative Study}
\label{sec:additional}
\subsection{Generating Longer Video Sequences}
\label{sec:long}
We slightly modify V3GAN to generate 32 frame sequences to investigate the generative capabilities of the network. We observe that even with longer sequences our model shows promising results. However, as expected the spatial as well as temporal quality is degraded. Ghost person (eighth row), person without head (second row), person with no action (seventh row), etc, appears frequently as shown in Figure \ref{fig:32frames}. The FID value 120.35 confirms the observed degradation in the quality of the generated videos.

\subsection{Conditional Video Generation}
\label{sec:conditional}
\begin{figure}
    \centering
    \includegraphics[scale=0.75]{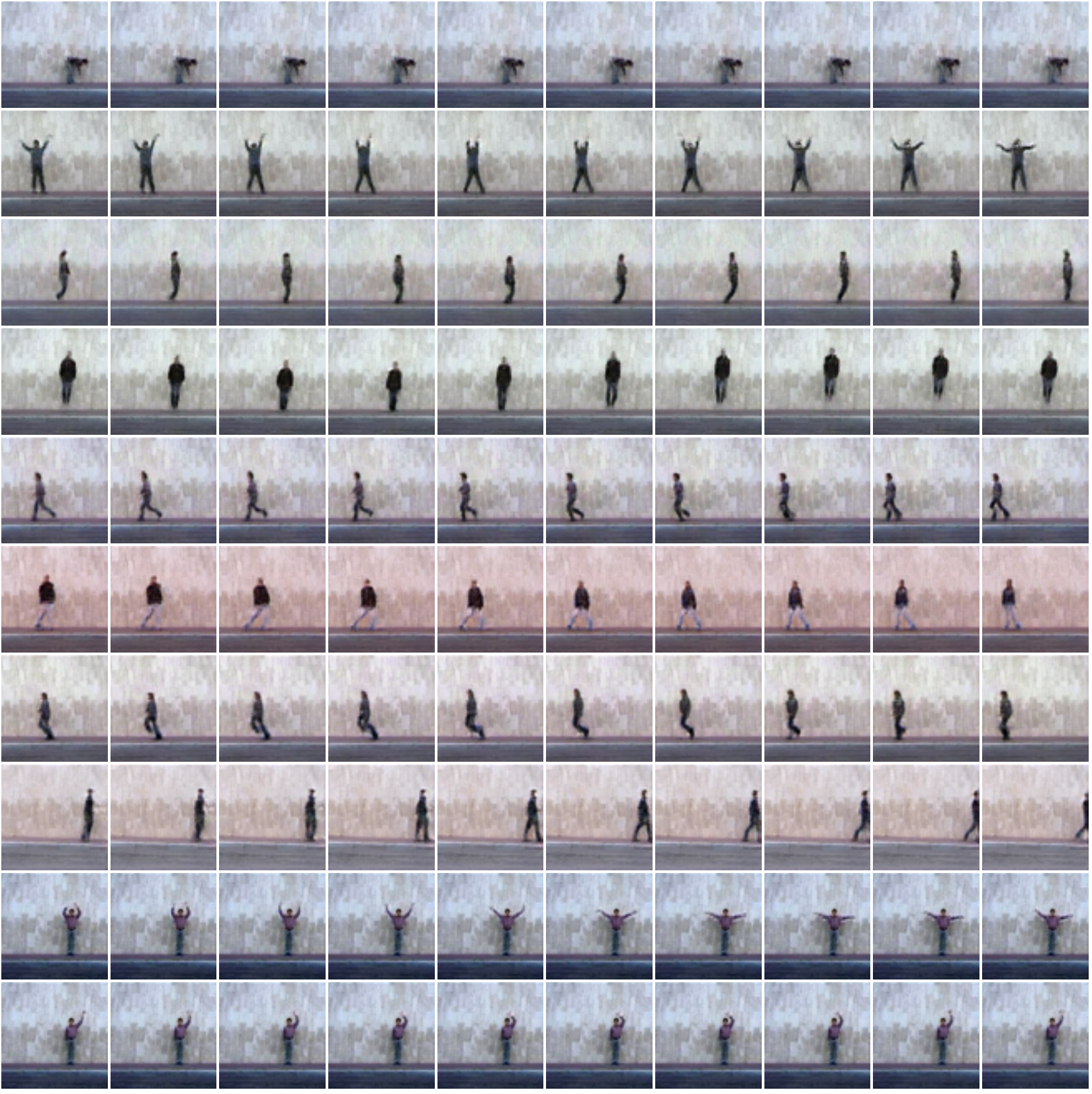}
    \caption{Qualitative results for Conditional video generation from V3GAN. Each row represents a video generated by combining randomly sampled $z_M$ and $z_{BG}$ with one hot vector representation of class label on Weizmann dataset. Class label used, top to bottom: bending, jumping jack,  jump, parallel jump, running, side, skip, walking, two hand waving, one hand waving.}
    \label{fig:conditional}
\end{figure}

We study the effect of the conditioning the model on class labels for Weizmann dataset. The class information is given to the generator by concatenating one hot vector representation of the category label with temporal noise $z_T$. For the discriminator, we reshape and repeat the class label representation  so that it can be concatenated as a channel with the input video. Similar approach has been followed for Image Discriminator. Figure \ref{fig:conditional} shows the results obtained by feeding random noise vectors with class information of Weizmann dataset to the generator. It can be seen that V3GAN can successfully generates the samples corresponding to each category of the dataset.  
 
\subsection{Latent Space Exploration}
\label{sec:latent_space}
\begin{figure}[H]
\centering
\includegraphics[scale=0.75]{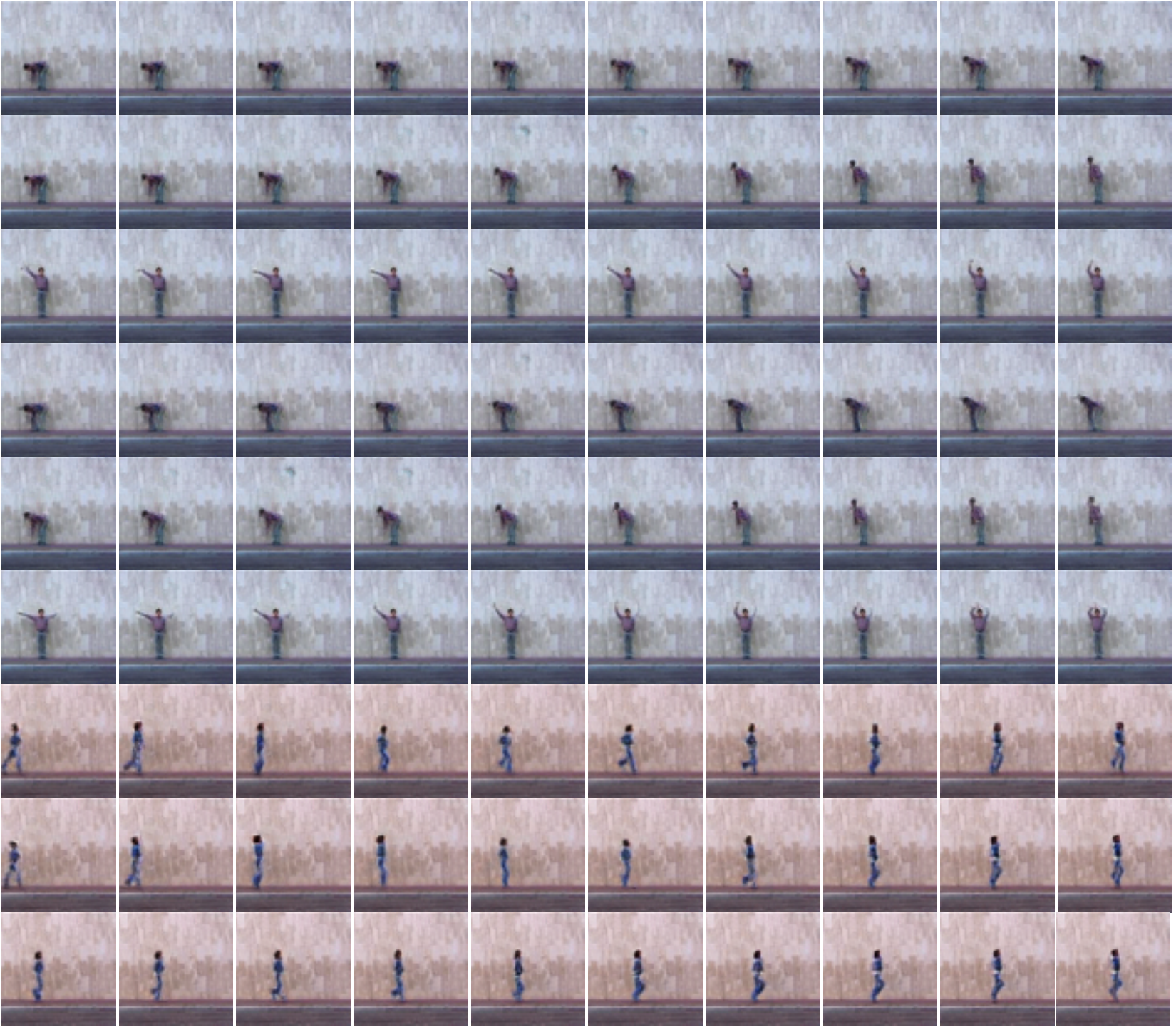}
 \caption{Generated samples for Latent space exploration on Weizmann dataset. We sample 3 different $z_{BG}$ and 3 $z_T$.  $z_{BG}$ is kept fixed for three consecutive rows, and $z_T$ is fixed for every $(3n+1)^{th}$ row, where $n \in {0,1,2}$. Each row represents a video sequence.}\vspace*{3cm}
\label{fig:wzm_latent_space2}
\end{figure}

\begin{figure}
    \centering
    \includegraphics[scale=0.7]{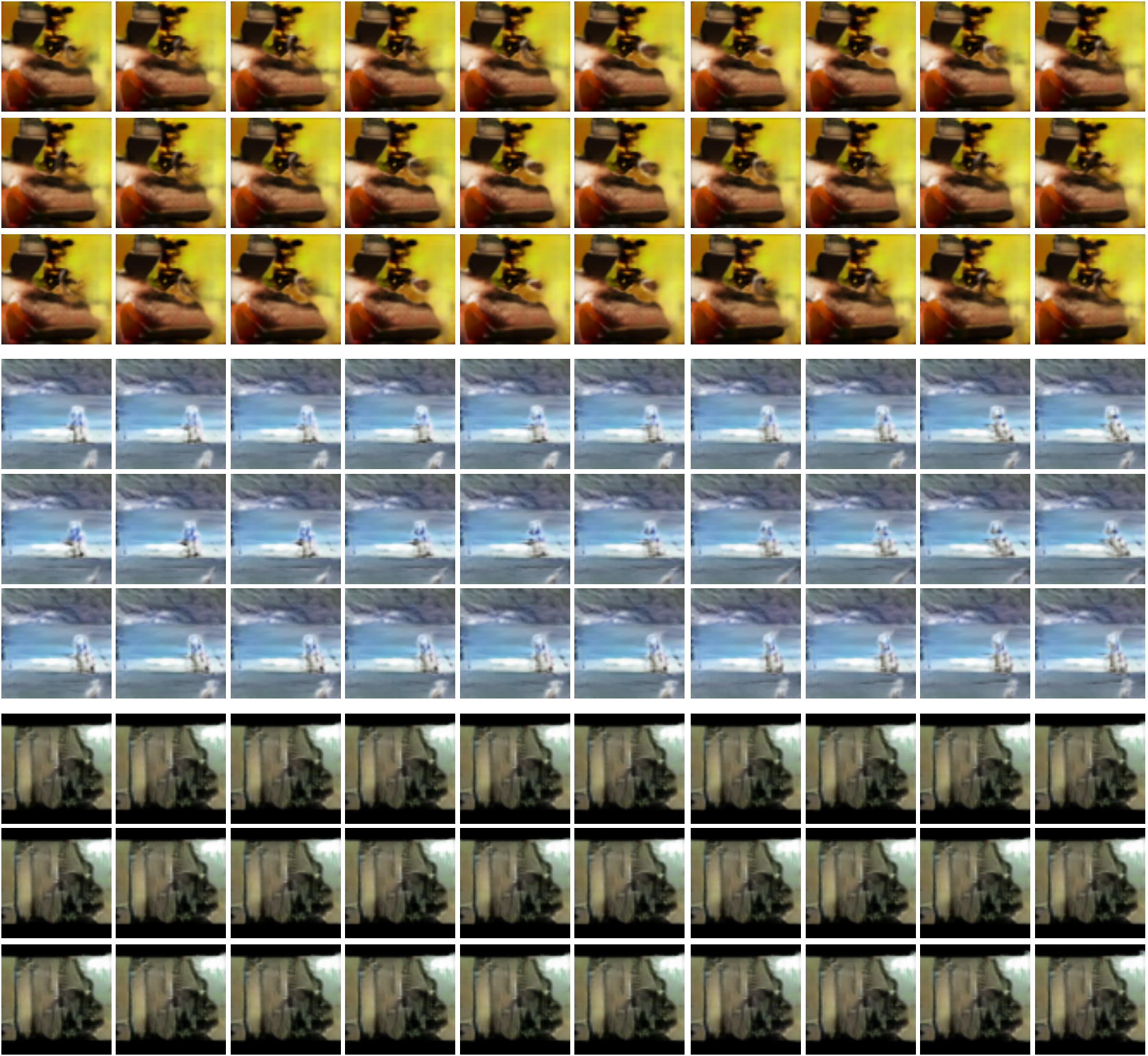}
    \caption{Generated samples for Latent space exploration on UCF101 dataset. We sample 3 different $z_{BG}$ and 3 $z_T$.  $z_{BG}$ is kept fixed for three consecutive rows, and $z_T$ is fixed for every $(3n+1)^{th}$ row, where $n \in {0,1,2}$. Each row represents a video sequence. First three rows seems to be Playing Daf, Next three rows shows surfing, last three rows corresponds to cliff diving.}
    \label{fig:ucf_noise_ablation}
\end{figure}

\begin{figure}[H]
\centering
\includegraphics[scale=0.75]{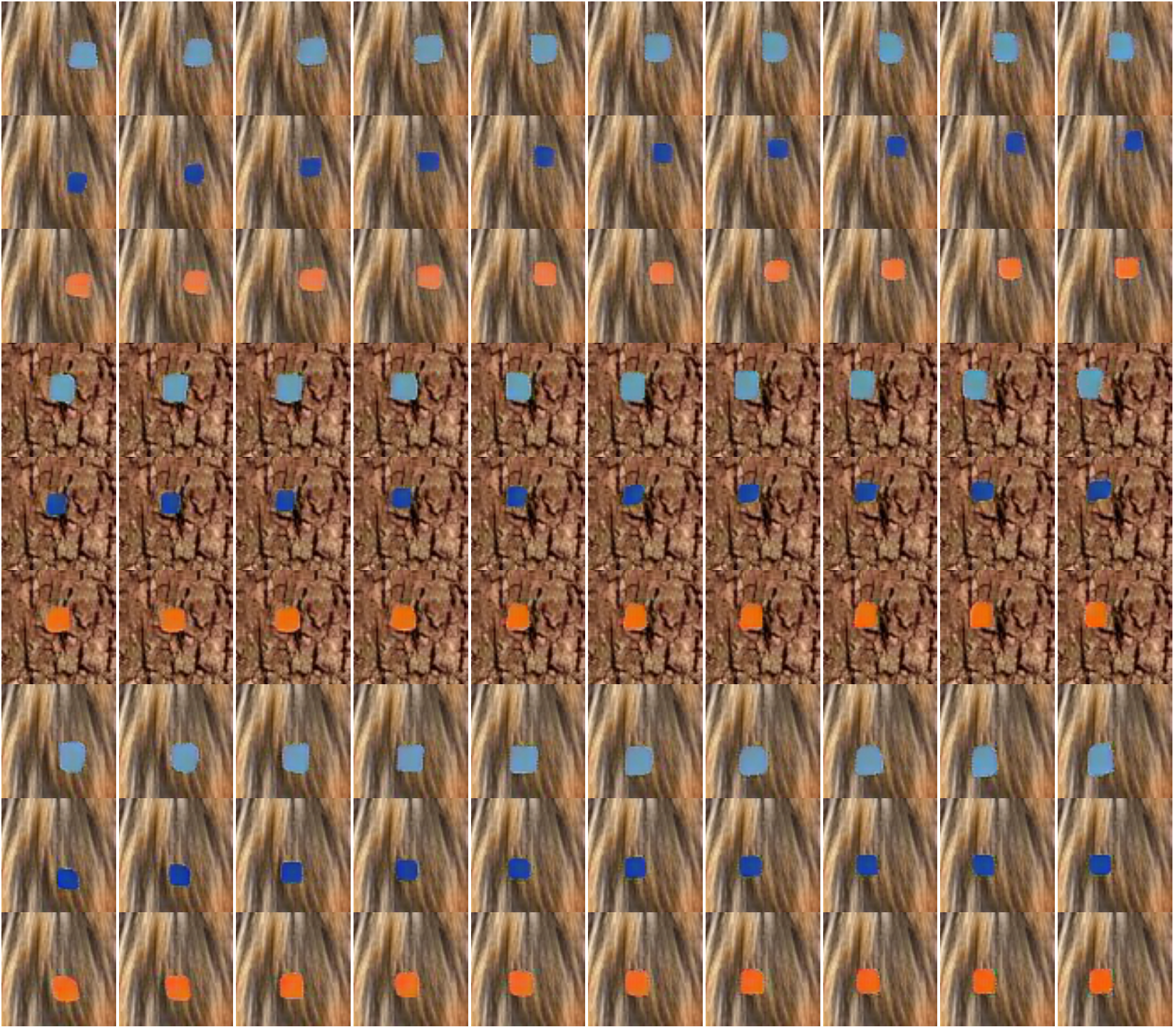}
\caption{Generated samples for Latent space exploration on Shapes dataset. We sample 3 different $z_{BG}$ and 3 $z_T$.  $z_{BG}$ is kept fixed for three consecutive rows, and $z_T$ is fixed for every $(3n+1)^{th}$ row, where $n \in {0,1,2}$. Each row represents a video sequence.}
\label{fig:shapes_latent_space}
\end{figure}

To understand the effect of latent vectors $z_{BG}$ and $z_T$, while generating videos, we use the following set-up. We sample a random noise vector $z_{BG}$ with three different $z_T$ and vice versa. We observe that for constant background noise, the background remain fixed and foreground depends on both $z_{BG}$ and $z_T$. Figure \ref{fig:wzm_latent_space2}, \ref{fig:ucf_noise_ablation},
\ref{fig:shapes_latent_space}, shows the latent space exploration for Weizmann, Shapes and UCF101 datasets, respectively. 

% \begin{figure}[H]
%     \centering

%  \caption{Latent space exploration for Shapes dataset. $z_{BG}$ is constant for each three consecutive rows, and $z_T$ is constant for every $(3n+1)^{th}$ row, where $n \in {0,1,2}$}
%     \label{fig:shapes_latent_space}
% \end{figure}

\section{Additional Ablation Study}
\label{sec:ablation}
\subsection{Architecture diagram for ablation studies}

\begin{figure}[H]
    \centering
    \includegraphics[width=\linewidth]{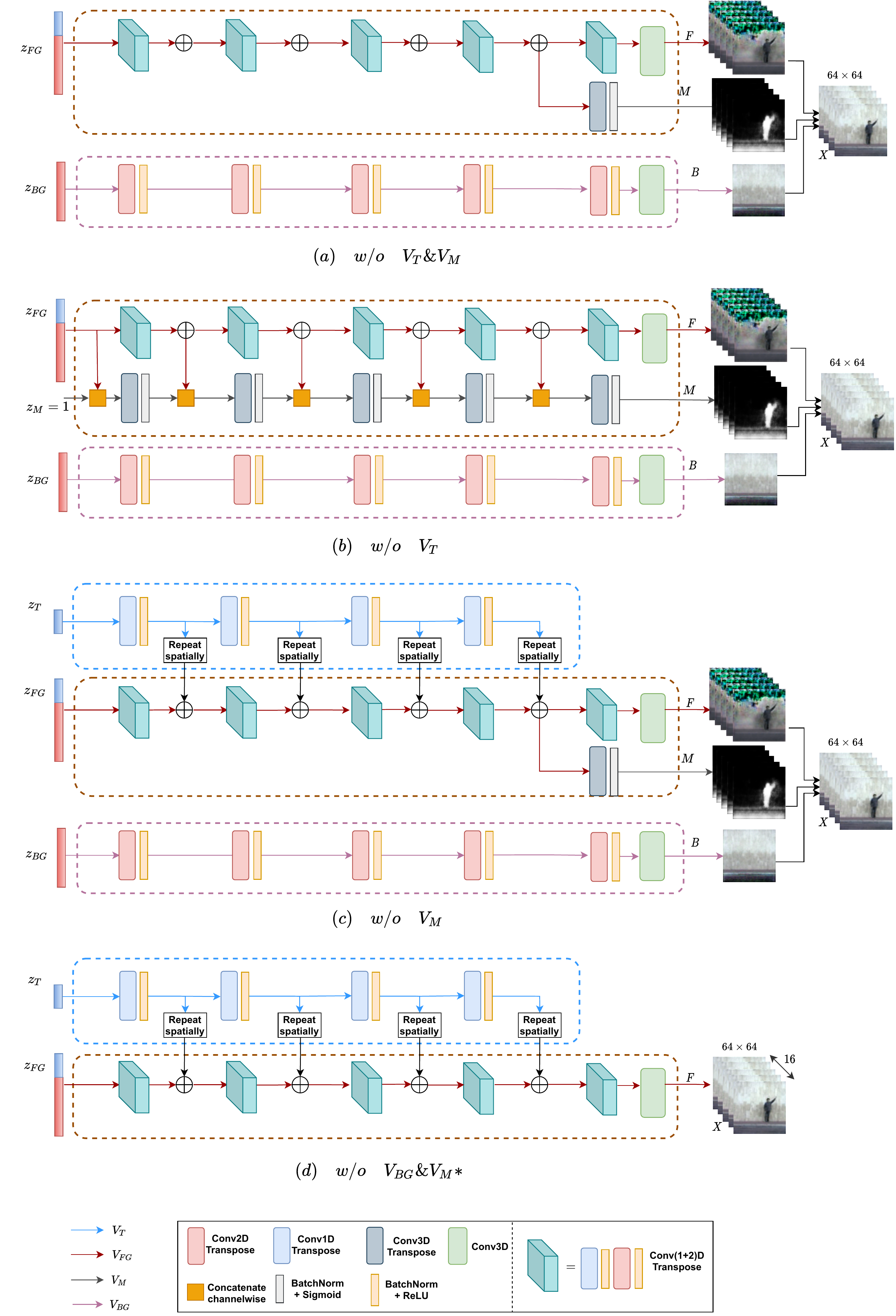}
    \caption{Architecture diagram for different ablation studies performed in the Main paper.}
    \label{fig:ablation}
\end{figure}

In figure \ref{fig:ablation}, we show the individual architectures for various ablations performed in Table 2 of the main paper. Notice how the last masking layer is retained in the configuration (a), (c) and how the entire masking  is removed in (d). Such a design is chosen so that we can explicitly examine the effect of the proposed feature level masking. 

\subsection{Effect of Gamma} 
To understand the weightage required for shuffling loss, we introduce the scaling parameter $\gamma$ in video discriminator as shown in equation \ref{video_descriminator_loss}. We train our network with different values of $\gamma$ like [10,1,0.1,0.25,0.5,0.75,0.01]. However, the best results are obtained for $\gamma = 1$. 
\begin{gather}
%F_I = E[\log{D_I(x_f)}]+E[\log(1-D_I(G(z_{BG},z_T)_f))] \\
F_{V} = E[\log{D_V(x)}]+E[\log(1-D_V(G(z_{BG},z_T)))] + \gamma L_{shuffle}(x)
\label{video_descriminator_loss}
\end{gather}

\section{More Qualitative Results}
\label{sec:qual}
% We provide additional samples generated from Shapes dataset in section \ref{sec:qual_shapes} and on UCF101 dataset in section \ref{sec:qual_ucf} using V3GAN. 

\subsection{Shapes Dataset and Generated samples on Shapes Dataset}
\label{sec:qual_shapes}
\begin{figure}[H]
    \centering
\includegraphics[scale=0.6]{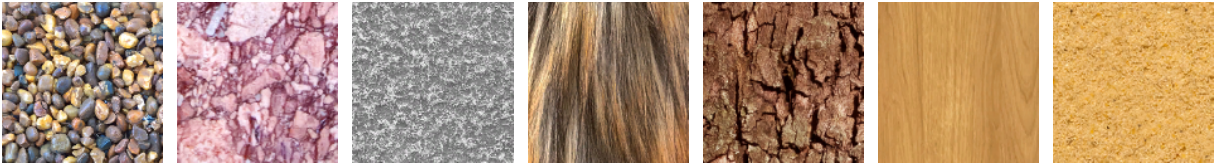}
 \caption{Seven background textures chosen for generating Shapes dataset.}
    \label{fig:shapes_bg}
\end{figure}

\begin{figure}[H]
    \centering
    \includegraphics[scale=0.6]{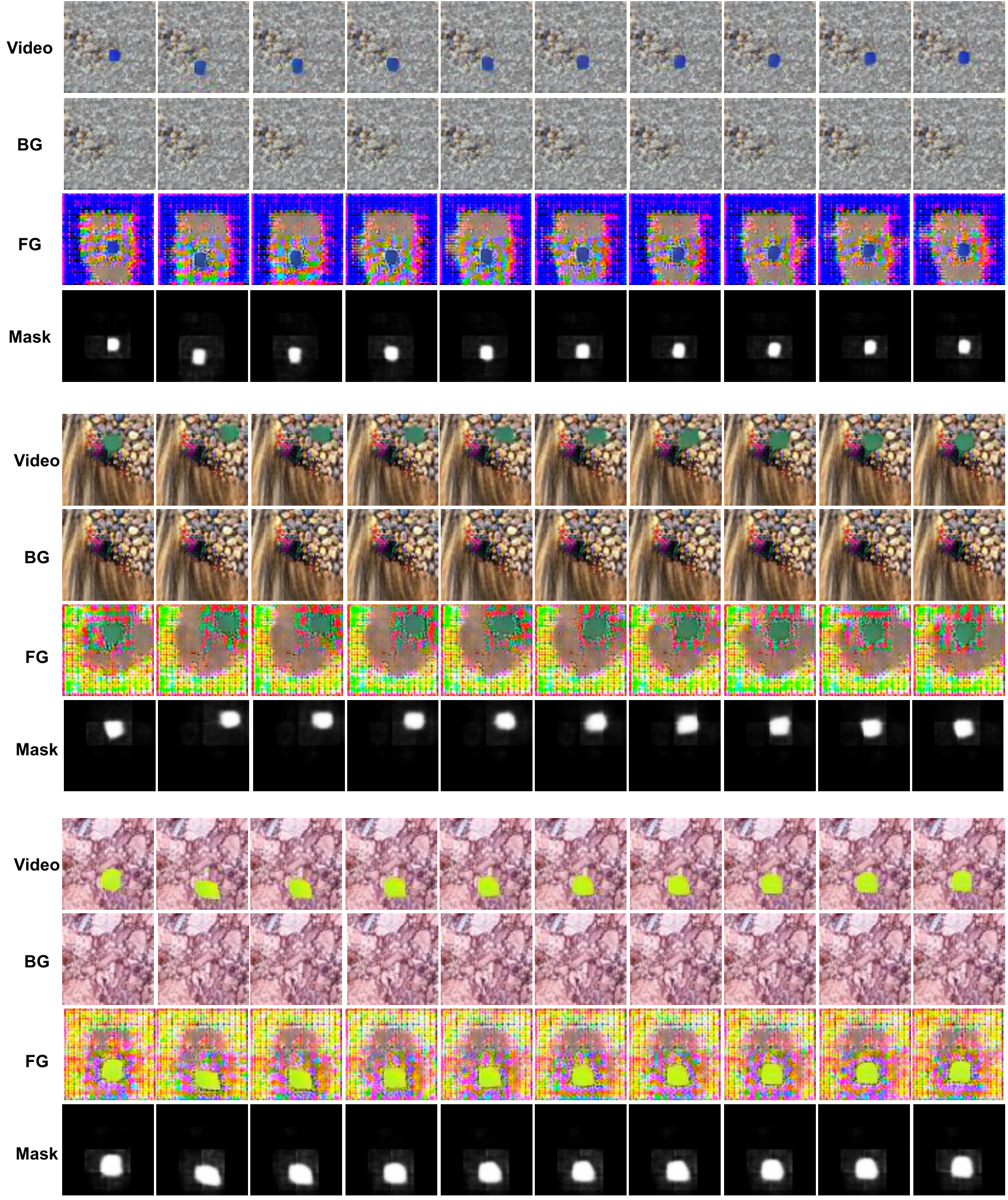}
    \caption{Generated video samples for Shapes dataset, along with corresponding background, foreground and mask generated from V3GAN.}
    \label{fig:shapes_samples}
\end{figure}

For Shapes dataset, we added a randomly sampled background out of seven textured backgrounds, for each video. Those backgrounds are shown in figure \ref{fig:shapes_bg}. The samples generated from Shapes dataset are shown in figure \ref{fig:shapes_samples}. 

\subsection{Generated Samples for UCF101 Dataset}
\label{sec:qual_ucf}
\begin{figure}[H]
    \centering
    \includegraphics[scale=0.7]{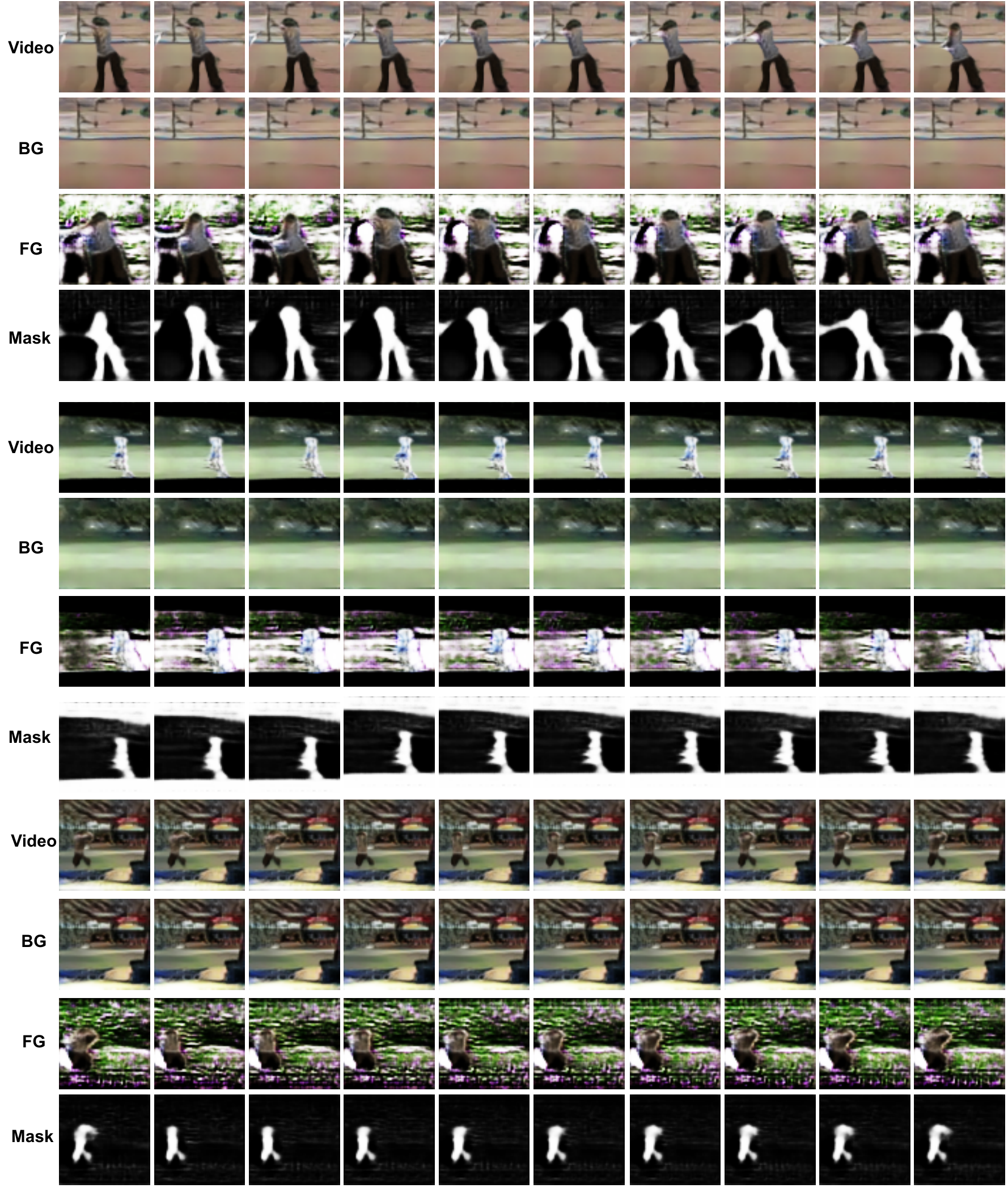}
    \caption{Generated video samples for UCF101 dataset, along with corresponding background, foreground and mask generated from V3GAN.}
    \label{fig:ucf_samples}
\end{figure}

\subsection{Some Failure Cases for Weizmann Dataset}
\begin{figure}
    \centering
    \includegraphics[scale=0.48]{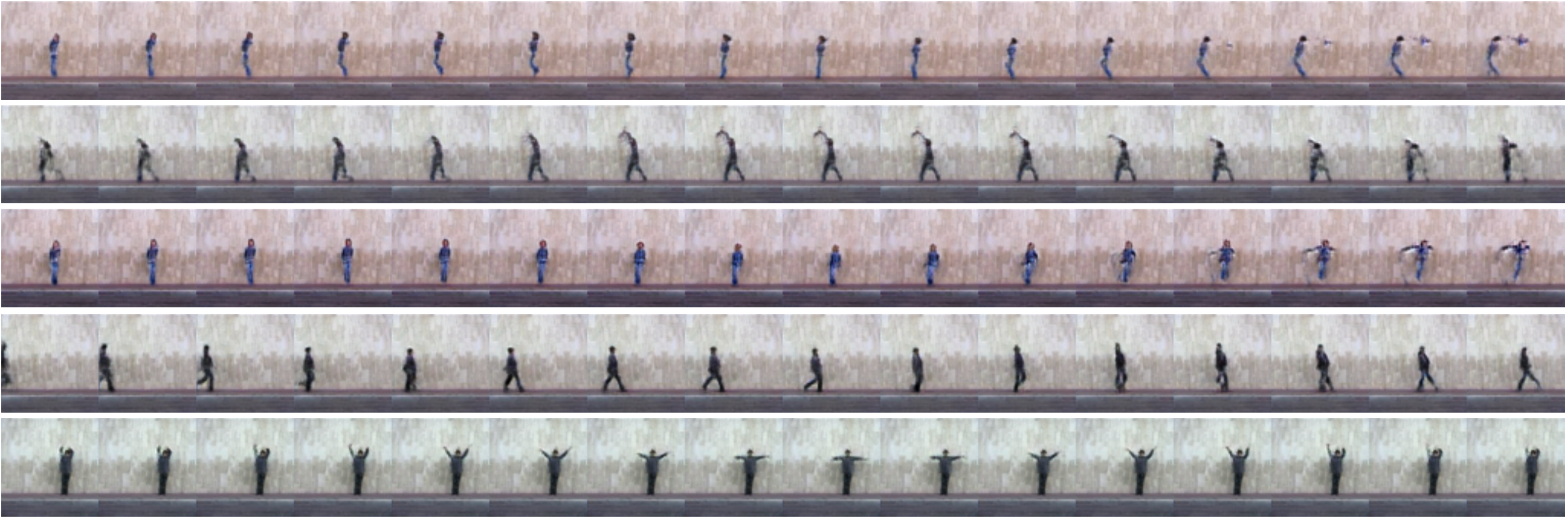}
    \caption{Failure cases in Weizmann Dataset.}
    \label{fig:fail}
\end{figure}